\documentclass{bmvc2k}
\usepackage{array}
\usepackage{amsfonts}

\DeclareMathOperator*{\argmax}{arg\,max}

\title{Deep Metric Learning Meets Deep Clustering: An Novel Unsupervised Approach for Feature Embedding~\footnote{This work was partially supported by the Australian Research Council project FT190100525.}}

\addauthor{Binh X. Nguyen}{binh.xuan.nguyen@aioz.io}{1}
\addauthor{Binh D. Nguyen}{binh.duc.nguyen@aioz.io}{1}
\addauthor{Gustavo Carneiro}{gustavo.carneiro@adelaide.edu.au}{3}
\addauthor{Erman Tjiputra}{erman.tjiputra@aioz.io}{1}
\addauthor{Quang D. Tran}{quang.tran@aioz.io}{1}
\addauthor{Thanh-Toan Do}{thanh-toan.do@liverpool.ac.uk}{2}

\addinstitution{
 AIOZ\\
 Singapore
}
\addinstitution{
 University of Liverpool\\
 United Kingdom
}

\addinstitution{
 University of Adelaide\\
 Australia
}

\runninghead{Binh X. Nguyen et al.}{Deep Metric Learning Meets Deep Clustering}

\begin{document}

\maketitle
%-------------------------------------------------------------------------
\begin{abstract}
{Unsupervised Deep Distance Metric Learning (UDML) aims to learn sample similarities in the embedding space from an unlabeled dataset.
Traditional UDML %Unsupervised Deep Distance Metric Learning (UDML)
methods usually use the triplet loss or pairwise loss which requires the mining of positive and negative samples w.r.t. anchor data points. This is, however, challenging in an unsupervised setting as the label information is not available. 
In this paper, we propose a new UDML method that overcomes that challenge. %uses deep clustering for deep metric learning. 
%Our method is a novel UDML method that  overcomes the challenge of not having the label information.
In particular, we propose to use a deep clustering loss to learn centroids, i.e., pseudo labels, that represent semantic classes. During learning, these centroids are also used to reconstruct the input samples. It hence ensures the representativeness of centroids --- each centroid represents visually similar samples. Therefore, the centroids give information about positive (visually similar) and negative (visually dissimilar) samples. Based on pseudo labels, we propose a novel unsupervised metric loss which enforces the positive concentration and negative separation of samples in the embedding space. %Our approach has $\mathcal{O}(KN)$ complexity --- linear in the number of centroids $K$ and the number of training images $N$.  
Experimental results on benchmarking datasets show that the proposed approach outperforms other UDML methods. % and is competitive to methods that are fully supervised.
%The source code is available at: \url{https://github.com/aioz-ai/BMVC20_CBSwR}
% we can publish the code, yet it is not mandatory to put the link in the paper. 
}
\end{abstract}

\section{Introduction}
{The objective of deep distance metric learning (DML) is to train a deep learning model that maps training samples into feature embeddings that are close together for samples that belong to the same category and far apart for samples from different categories \cite{dosovitskiy2014discriminative,han2015matchnet,smart,global,masci2014descriptor,pascal,Simonyanpami14,lifted,histogram,wohlhart2015learning,ZagoruykoCVPR15,shrivastava2016training}. %The use of DML is more flexible compared to the traditional classification models because DML does not impose strong constraints on the number of classes that the model can handle. Furthermore, DML requires no model structure update, which means that the learned DML model can simply be fine-tuned with the new dataset. Therefore, DML is an attractive approach to be used in learning problems that can be continuously updated, such as open-world~\cite{bendale2015towards} and life-long learning problems~\cite{thrun2012learning}.}
{Traditional DML approaches require supervised information, i.e., class labels, to supervise the training. The class labels are used either for training the pointwise softmax loss \cite{genericfea,genericfea0} or mining positive and negative samples for training the pairwise or triplet losses~\cite{siamac,pascal,contrastive0,facenet,smart,DBLP:conf/cvpr/QianJZL15}. 
Although the supervised DML achieves impressive results on different tasks \cite{shi2016embedding,wang2016joint,zhou2017efficient,chen2017beyond,facenet,hu2014discriminative}, it requires large amount of annotated training samples to train the model. Unfortunately, such large datasets are not always available and they are costly to annotate for specific domains. That disadvantage also limits the transferability of supervised DML to new domain/applications which do not have labeled data. 
These reasons have motivated recent studies aiming at learning feature embeddings without annotated datasets~\cite{lu2015surpassing,tripletproxy,ye2019unsupervised} --- unsupervised deep distance metric learning (UDML). Our study is in that same direction, i.e., learning embeddings from unlabeled data. 
}

{There are two main challenges for UDML. Firstly, how to define positive and negative samples for a given anchor data point, such that we can apply distance-based losses, e.g., pairwise loss or triplet loss, in the embedding space. Secondly, how to make the training efficient, given a large number of pairs or triplets of samples, in the order of $\mathcal{O}(N^2)$ or $\mathcal{O}(N^3)$, respectively, in which $N$ is the number of training samples.
}
In this paper, we propose a new method that utilizes deep clustering for deep metric learning to address the two challenges mentioned above.

Regarding the first challenge, given an anchor point, its transformed versions (obtained, for example, from data augmentation methods) can be used as positive samples~\cite{rotation-iclr18,ye2019unsupervised}. Alternatively, the Euclidean nearest neighbors from pre-trained features can be used for selecting positive samples \cite{hadsell2006dimensionality}. The problem is more challenging for mining negative samples. In \cite{iscen2018mining}, the authors rely on manifold mining which uses the pre-trained features of samples to build a similarity graph and the mining is performed on the graph. Hence, this approach heavily depends on pre-trained features. %More specifically, the learning of embedding features is guided by the pre-trained features.
In addition, building a full adjacency matrix for graph mining is costly, especially for large scale training datasets.   
Recently, in \cite{ye2019unsupervised}, the authors consider all other training samples within the same batch of the anchor point as negative samples (w.r.t. the anchor) --- this means that no negative mining is performed at all. This approach likely contains false negatives, especially when the number of same class samples in a batch is large. Different from these approaches, we propose to learn pseudo labels for samples through a deep clustering loss. The pseudo labels are then used for negative mining.} 

{Regarding the second challenge, due to the large number of pairs or triplets w.r.t. all training samples, traditional approaches reduce the space by only considering $\mathcal{O}(m^2)$ pairs or $\mathcal{O}(m^3)$ triplets within each training batch containing $m$ samples. %, making the asymptotic complexity for training for the whole dataset $\mathcal{O}(Nm)$ or $\mathcal{O}(Nm^2)$, respectively. However, even with this reduced space, the number of pairs/triplets is still large when training with a large batch size. 
In our approach, the reliance on the pseudo labels allows us to leverage the idea of center loss \cite{DBLP:conf/eccv/WenZL016,do2019theoretically} which results in a training complexity 
of $\mathcal{O}(Km)$ for each batch, in which $K$ is the number of the learned centroids (i.e., pseudo labels). %In other words, the training complexity for the whole dataset is $\mathcal{O}(NK)$ in which $K$ is usually smaller than $m$. 
%NOTE FOR US: but because our network has a decoder part, the real training time is not faster than CVPR'19 (O(Nm)).
}

Our main contributions can be summarized as follows. (i) We propose an novel UDML approach with a novel loss function that consists of a deep clustering loss for learning pseudo class labels, i.e., clusters, where each cluster is represented by a centroid. 
(ii) To enhance the representativeness of centroids, the loss also contains a reconstruction loss term that penalizes high sample reconstruction errors from centroid representations, which will encourage visually similar samples to belong to the same cluster. (iii) Based on pseudo labels, we propose a novel center-based loss that encourages samples to be close to their augmented versions and far from the centroids from different clusters in the embedding space. %The proposed method is evaluated on benchmarking datasets and the experimental results show that our method is outperforming the state-of-the-art UDML methods and is competitive to other fully supervised DML methods. 

\section{Related Work}
%{This section presents a brief overview of related works on supervised deep metric learning, unsupervised feature learning, and unsupervised deep metric learning.}
\paragraph{Supervised deep metric learning.} 
{The supervised deep metric learning  uses the label information to supervise training \cite{smart,tripletproxy,facenet,pascal,deng2019arcface,ranjan2019l2,wang2018additive,wang2018cosface,hadsell2006dimensionality,oh2016deep,npair,wang2019ranked,wang2019multi}. Generally, the common loss functions used in supervised metric learning can be divided into two types: classification-based losses and distance-based losses. The classification-based losses focus on maximizing the probability of correct labels predicted for every sample \cite{deng2019arcface,ranjan2019l2,wang2018additive,wang2018cosface}. Those methods have linear run-time training complexity $\mathcal{O}(NC)$ where $N$ and $C < N$ represent the number of training samples and the number of classes, respectively. The distance-based losses \cite{hadsell2006dimensionality,oh2016deep,facenet,npair,wang2019ranked,wang2019multi} focus on learning the similarity or dissimilarity among sample embeddings. Two common distance-based losses are the contrastive (pairwise) loss \cite{hadsell2006dimensionality} and the triplet loss \cite{hoffer2015deep,do2019theoretically}. The run-time training complexity of contrastive loss is $\mathcal{O}(N^2)$ while triplet loss is $\mathcal{O}(N^3)$, in which $N$ is the number of training samples. Hence both losses suffer from slow convergence due to a large number of trivial pairs or triplets as the model improves.}

{In order to overcome the run-time complexity challenge, many studies focused on mining strategies aiming at finding informative pairs or triplets for the training \cite{smart,tripletproxy,facenet,pascal}. For example, in~\cite{smart}, the authors propose to use fast approximate nearest neighbor search for quickly identifying informative (hard) triplets for training, reducing the training complexity to $\mathcal{O}(N^2)$. In~\cite{tripletproxy}, the mining is replaced by the use of $P < N$ proxies, where a triplet is re-defined to be an anchor point, a similar and a dissimilar proxy -- this reduces the training complexity to $\mathcal{O}(NP^2)$.
Recently, in \cite{do2019theoretically} the authors propose a center-based metric loss in which centers are fixed during training and the loss involves the distance calculations between feature embeddings and the centers, rather than between features, resulting in  $\mathcal{O}(NC)$ complexity, in which $C$ is the number of centers.}
\paragraph{Unsupervised feature learning.}  
{The common approach to unsupervised feature learning is to learn intermediate features, i.e., latent features, that well represent  the input samples. %Representative methods are based on generative models, such as Auto-encoder \cite{bengio2007scaling,bengio2009learning} and GAN \cite{DBLP:conf/nips/GoodfellowPMXWOCB14}. 
Representative works are Autoencoder-based methods~\cite{vincent2008extracting,rifai2011contractive}. 
Another approach is to leverage the idea of clustering, such as in~\cite{krause2010discriminative,hu2017learning}, where features are learned with the objective that they can be clustered into balanced clusters and the clusters are well separated. In \cite{caron2018deep}, the authors propose to jointly learn the parameters of a deep neural network and the cluster assignments of the resulting features.  More precisely, the method iterates between the k-means clustering of the features produced by the network and the updating of network weights by predicting the cluster assignments as pseudo labels using a discriminative loss.
%\gustavo{One issue with deep clustering~\cite{caron2018deep} is that the clustering happens only at the beginning of the epoch, and is not updated during the training process, which is different from our approach that continuously updates the clustering.  Another issue is that deep clustering only uses the pseudo labels for supervising a classification loss and not the cluster centers, which can potentially limit the potential of the training process -- our method uses both the pseudo labels and cluster centers.}
}

{Another popular approach to unsupervised feature learning is to replace the labels annotated by humans by labels generated from the original data with data augmentation methods.  %or by labels generated from geometry information. 
For example, in \cite{dosovitskiy2015discriminative} the authors train a deep network to discriminate between a set of surrogate classes. Each surrogate class is formed by applying a variety of transformations to a randomly sampled image patch. In \cite{DBLP:conf/cvpr/LinLCZ16}, the authors train a deep network such that it minimizes the distance between the features that describe a reference image and their rotated versions. Similarly, in  \cite{rotation-iclr18}, the authors train a convnet to predict the rotation that is applied to the input image.
%Regarding creating labels from geometry information, in \cite{pascal} the stereo correspondence image patches are used as supervision information to train a Siamese network. In \cite{siamac,DBLP:conf/wacv/DoHTPLC019} 3D reconstructed models are used for creating the supervision information, i.e., images belonging to the same 3D model that share enough 3D points are considered as positive pairs, while images belonging to different 3D models are considered as negative pairs.
}
\paragraph{Unsupervised deep metric learning.}
{The unsupervised deep metric learning can be considered as a special case of unsupervised feature learning. In this case, the networks are trained to produce deep embedding features from unlabeled data. Furthermore, the learned models are expected to generalize well to unseen testing class samples. Regarding this problem, there are only few works proposed to tackle the problem. In \cite{iscen2018mining} the authors propose a mining strategy using Euclidean neighbors and manifold neighbors to pre-mine positive and negative pairs. The mining process involves building an Euclidean nearest neighbor graph from pre-computed features, a process that has $\mathcal{O}(N^2)$ complexity, in which $N$ is the number of training samples. This means that the approach is not scalable and also heavily depends on the pre-computed features. 
To overcome the mining complexity, in \cite{ye2019unsupervised}, the authors propose a softmax-based embedding which enforces data augmentation invariance. Specifically, for each sample in a batch, its augmentations are considered as the positive samples and other samples are treated as negative ones. %Therefore, the approach likely creates the false negative samples.
During training, the method needs to calculate the pairwise distance between samples in a batch, so its complexity for training a batch is quadratic w.r.t. batch size $m$. %Hence the asymptotic complexity for training the whole dataset is $\mathcal{O}(Nm)$. 
}

%{\gustavo{There is a comment in the reviews that I don't see addressed here, which is about the key differences and advantages of the proposed method over these self-supervised learning methods.}}

\section{Method}
\begin{figure}
    \centering
    \includegraphics[width=0.8\textwidth, keepaspectratio=true]{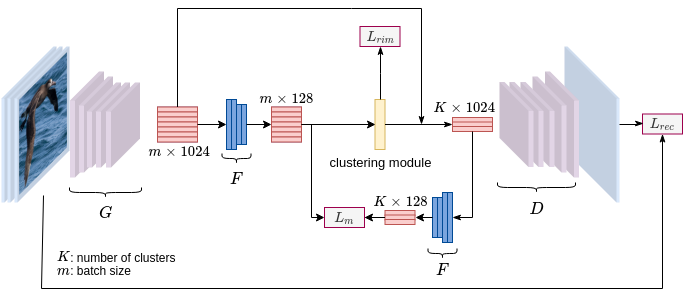}
    \caption{\small Illustration of the proposed framework which consists of an encoder (G), an embedding module (F), a decoder (D) and three losses, i.e., clustering loss $L_{rim}$, reconstruction loss $L_{rec}$ and metric loss $L_m$. The details are presented in the text.  
    %Input images are fed into the backbone network which is also considered as the encoder (G -- purple) to get image representations (red) which have 1024 dimensions. The image representations are passed through the embedding module which consists of fully connected layers and $L2$ normalization layers (F -- blue), which results image embeddings with 128 dimensions. Clustering module (yellow) takes image embeddings as inputs and performs the clustering with a clustering loss ($L_{rim}$) and outputs the cluster assignments. Given the cluster assignments, centroid representations are computed from image representations,  then are passed through the decoder with a reconstruction loss ($L_{rec}$) to reconstruct images that belong to the corresponding clusters. The centroid representations are also passed through the embedding module (F) to get centroid embeddings. The centroid embeddings and image embeddings are used as inputs for the metric loss ($L_m$).
    }
    \label{fig:framework}
\end{figure}
{
The proposed framework is presented in Figure \ref{fig:framework}. For every original image in a batch, we make an augmented version by using a random geometric transformation. Let the number of input images in a batch after augmentation be $m$. The input images are fed into the backbone network which is also considered as the encoder (G -- purple) to get image representations (red) which have 1024 dimensions. The image representations are passed through the embedding module which consists of fully connected and $L2$ normalization layers (F -- blue), which results in {unit norm} image embeddings with 128 dimensions. The clustering module (yellow) takes image embeddings as inputs, performs the clustering with a clustering loss ($L_{rim}$), and outputs the cluster assignments. Given the cluster assignments, centroid representations are computed from image representations, which are then passed through the decoder with a reconstruction loss ($L_{rec}$) to reconstruct images that belong to the corresponding clusters. The centroid representations are also passed through the embedding module (F) to get centroid embeddings. The centroid embeddings and image embeddings are used as inputs for the metric loss ($L_m$).
}

\subsection{Discriminative Clustering}
\label{sec3:cl_loss}

{
Following the idea of Regularized Information  Maximization (RIM) for clustering  \cite{krause2010discriminative,hu2017learning}, we formulate the clustering of embedding features as a classification problem. Given a set of embedding features 
$X=\{x_i\}_{i=1}^m \in \mathbb{R}^{128 \times m}$ in a batch and the number of clusters $K\le m$ {(i.e., the number of clusters $K$ is limited by the batch size $m$)}, 
we want to learn a probabilistic classifier, using a softmax activation, that maps each $x \in X$ into a probabilistic vector $y=softmax_{\theta}(x)$ %$p_{\theta}(y|x)$ 
that has $K$ dimensions, in which $\theta$ is the classifier parameters\footnote{In this context, the softmax classifier consists of an FC layer parameterized by $\theta$ that projects the feature embedding $x$ to a vector with $K$ dimensions, i.e., the logits. Then the softmax function is applied on the logits to output a probabilistic vector.
%https://stackoverflow.com/questions/34240703/what-is-logits-softmax-and-softmax-cross-entropy-with-logits
}. The cluster assignment for $x$ then is estimated by %$c^* = \argmax_c p_{\theta}(y|x)$
$c^* = \argmax_c y$. 

Let %$Y \in \RR^{K\times m}$ 
$Y=\{y_i\}_{i=1}^m$ be the set of softmax outputs for $X$. Inspired by RIM \cite{krause2010discriminative,hu2017learning}, we use the following  objective function for the clustering
\begin{equation}
    L_{rim} = \mathcal{R}(\theta) - \lambda \left[ H(Y) - H(Y|X) \right],
    \label{eq:c_loss}
\end{equation}
where $H(.)$ and $H(.|.)$ are entropy and conditional entropy, respectively; $\mathcal{R}(\theta)$ regularizes the classifier parameters (in this work we use $l_2$ regularization); $\lambda$ is a weighting factor to control the importance of two terms. 
%The term $[H(Y) - H(Y|X)]$ is actually the mutual information between $X$ and $Y$. By minimizing (\ref{eq:c_loss}), we maximize the mutual information between $X$ and $Y$. 
%Specifically, 
Minimizing (\ref{eq:c_loss}) is equivalent to maximizing $H(Y)$ and minimizing $H(Y|X)$.  Increasing the marginal entropy $H(Y)$ encourages cluster balancing, while decreasing the conditional entropy $H(Y|X)$ encourages cluster separation \cite{bridle1992unsupervised}.  Following \cite{hu2017learning} the entropy and the conditional entropy are calculated over the batch as follows
\begin{equation}
    H(Y) = h\left( \frac{1}{m}\sum_{i=1}^m y_i \right),
    \label{eq:entropy}
\end{equation}
\begin{equation}
    H(Y|X) = \frac{1}{m} \sum_{i=1}^m h(y_i),
    \label{eq:con_entropy}
\end{equation}
where $h(y_i) \equiv -\sum_{j=1}^K y_{i_j} \log y_{i_j}$ is the entropy function.
}

\subsection{Reconstruction}
\label{sec3:r_loss}
{
In order to enhance the representativeness of centroids, we introduce a reconstruction loss that penalizes high reconstruction errors from centroids to corresponding samples. Specifically, the decoder takes a centroid representation of a cluster and minimizes the difference between input images that belong to the cluster and the reconstructed image from the centroid representation. By jointly training both the clustering loss and the reconstruction loss, we expect that centroids will well represent the samples belonging to the corresponding clusters. %In other words, visually similar samples are expected to be grouped into the same clusters. 
%The reconstruction criterion aims to minimize the difference between the input images and the reconstruction of centroid representation. The objective function could be seen as a regularization to clustering loss. 

Let $X_j$ be a set of input images in the batch that belongs to the cluster $j$. The centroid representation for the cluster $j$ is calculated as 
\begin{equation}
    r_j = \frac{1}{|X_j|}\sum_{I_i \in  X_j}G(I_i),
    \label{eq:centroid}
\end{equation}
where $G(.)$ is the backbone network (i.e., the encoder), which extracts image representations.
}
{
After obtaining the centroid representations, the reconstruction loss function is calculated as
\begin{equation}
    L_{rec} = \frac{1}{m} \sum_{j=1}^K\sum\limits_{I_i\in X_j}||I_i - D(r_j)||^2,
    \label{eq:r_loss}
\end{equation}
where $D(.)$ is the decoder which reconstructs samples in the batch using their corresponding centroid representations and $m$ is the number of images in the batch. 

%It is worth noting that we use the outputs of the encoder (that have $1024$ dims) rather than the outputs of the embedding layer (that have $128$ dims) as the centroid representations which are used as inputs for the decoder. We empirically found that this setting boosts the performance and produces better reconstructed images. 
%\gustavo{Even though the reconstructed image will not look similar from a visual perception viewpoint, as we will show in Sec.~\ref{sec:experiments} (Fig.~\ref{fig:reconstruction}), it will enforce the learning of representative centroids that capture the essential visual information of the images, also to be shown in Sec.~\ref{sec:experiments} (Fig.~\ref{fig:reconstructedCUB}).}
}

\subsection{Metric Loss}
\label{sec3:ml_loss}
{
For every original image $I_i$ in a batch, let $\hat{I_i}$ be its augmented version. Let $f_i \in \mathbb{R}^{128}$ and $\hat{f_i} \in \mathbb{R}^{128}$ be the image embeddings of $I_i$ and $\hat{I_i}$, respectively. 
The proposed metric loss aims to minimize the distance between $f_i$ and $\hat{f_i}$ while pushing $f_i$ far away from negative clusters\footnote{Hereafter, $I_i$ and $\hat{I_i}$ are interchangeable when calculating losses.}.  

Given the cluster representation $r_j$ for cluster $j$ (see Section \ref{sec3:r_loss}), the centroid embedding $c_j$ is obtained by $c_j = F(r_j) \in \mathbb{R}^{128}$, where $F(.)$ is the embedding module. It is worth noting that both the image embedding $f$ and centroid embedding $c$ are unit norm. Let the index of the cluster that sample $i$ belongs to be $q$, $1 \le q \le K$. 
In order to minimize the distance between  $f_i$ and $\hat{f_i}$, and maximize the distance between $f_i$ and centroids that $I_i$ does not belong to, we maximize the following ``softmax'' like function
}
{
\begin{equation}
    %P(i|\hat{x}_i) = \frac{\exp(f_i^T {\hat f_i}/\tau)}{\sum_{k=1}^K \exp(f_i^T c_k/\tau)}
    l(I_i, \hat{I_i}) = \frac{\exp(f_i^T {\hat f_i}/\tau)}{\sum_{k=1,k \neq q}^K \exp(f_i^T c_k/\tau)},
    \label{eq:close}
\end{equation} %max product (f_i,fi^) <--> min Euclidean (f_i,fi^). 
where $\tau$ is the temperature parameter \cite{DBLP:journals/corr/HintonVD15}. Using a higher value for $\tau$ produces a softer probability distribution. 

Because feature embeddings and centroid embeddings are unit norm, maximizing (\ref{eq:close}) minimizes the Euclidean distance between $f_i$ and $\hat{f_i}$ and maximizes the Euclidean distances between $f_i$ and the clusters that $I_i$ does not belong to. 
}

{
In addition, to push $f_i$ far away from a cluster $j$ that $I_i$ does not belong to, we maximize the following quantity, $\forall j \neq q$
\begin{equation}
    l(I_i,c_j) = 1 - \frac{\exp(f_i^T c_j/\tau)}{\sum_{k=1}^K \exp(f_i^T c_k/\tau)}.
    \label{eq:far}
\end{equation} % <--> max Euclidean distance between (f_i, c_j) <--> min product (f_i,c_j) <---> max (1 - product (f_i,c_j))
By maximizing (\ref{eq:far}) $\forall j \neq q$, we maximize the Euclidean distance between $f_i$ and $c_j$ while minimizing the Euclidean distance between $f_i$ and the centroid that $I_i$ belongs to, i.e., $c_q$.
%Under the assumption that $l(I_i,c_j)$ quantities are independent for different $j = 1,...,K$. In order to minimize the distance between $f_i$ and $\hat{f_i}$, and maximize the distance between $f_i$ and clusters in which $I_i$ does not belong to,
Combining (\ref{eq:close}) and (\ref{eq:far}), we maximize the following quantity for each sample $i$
\begin{equation}
    l(i) = l(I_i, \hat{I_i}) \prod\limits_{j=1,j\neq q}^{K}l(I_i, c_j).
    \label{eq:join_prob}
\end{equation}

Applying the negative log-likelihood to \eqref{eq:join_prob} and summing over all training samples in the batch, we minimize the metric loss function which is defined as follows
}
{
\begin{equation}
    L_{m} = -\sum_i log \left( l(I_i,\hat{I_i}) \right) - \sum_i\sum\limits_{j=1, j \neq q}^K log \left(l(I_i,c_j) \right).
    \label{eq:ml_loss}
\end{equation}
}
%\gustavo{Notice that only centroids are used as negative points in the metric loss of~\eqref{eq:ml_loss}, which can be a potential problem if the data set contains large variations. However, given that clustering is performed per batch, we note that such variation is relatively small, which means that important data set information is kept during the training process.}

\subsection{Final Loss}
{
The network in Figure \ref{fig:framework} is trained in an end-to-end manner with the  following multi-task loss
\begin{equation}
    L = \alpha L_{m} + \beta L_{rim} + \gamma L_{rec},
    \label{eq:final_loss}
\end{equation}
where $L_m$ is the center-based softmax loss for deep metric learning (\ref{eq:ml_loss}), $L_{rim}$ is the clustering loss (\ref{eq:c_loss}), and $L_{rec}$ is the reconstruction loss (\ref{eq:r_loss}).

\paragraph{Complexity.} In our metric loss $L_{m}$, we calculate the distance between samples and centroids, resulting in a $\mathcal{O}(Km)$ complexity. Both clustering loss $L_{rim}$ and reconstruction loss $L_{rec}$ have $\mathcal{O}(m)$ complexity. Hence the overall asymptotic complexity for training one batch of the proposed method is $\mathcal{O}(Km)$.

%\paragraph{Comparison with Softmax Embedding \cite{ye2019unsupervised}.} It is worth noting that the proposed metric loss (\ref{eq:ml_loss}) has  similar form to the Softmax Embedding loss --- SME in \cite{ye2019unsupervised}. However, in SME, there are no centroids involved. For each image in the batch, all other images but the augmented version, are considered as the negative samples. The distances in (\ref{eq:close}) and (\ref{eq:far}) are calculated between pair of samples, so the training complexity for a batch is $\mathcal{O}(m^2)$. In our metric loss, we calculate the distance between samples and centroids, resulting in a $\mathcal{O}(Km)$ complexity. This is particularly advantageous when training with large batch sizes. In addition, our final loss (\ref{eq:final_loss}) also consists of a clustering loss and a reconstruction loss for learning good centroid representations, i.e., pseudo labels. This allows us to perform more accurate negative mining than \cite{ye2019unsupervised}. In our method, both clustering and reconstruction losses have $\mathcal{O}(m)$ complexity. Hence the overall asymptotic complexity for training one batch of the proposed method is $\mathcal{O}(Km)$. \gustavo{Note that the run-time complexity of our method is favourable for large batch-sizes, particularly when $m >> K$.}
}

\section{Experiments}
\label{sec:experiments}
\subsection{Implementation Details}
{%The proposed method is implemented with  Pytorch. 
Following existing methods \cite{iscen2018mining,ye2019unsupervised,tripletproxy,oh2016deep,npair}, the pre-trained GoogLeNet %Inception-V1 
\cite{szegedy2015going} on ImageNet is used as the backbone network (G). %As standardly done in the literature, after the last average pooling of the GoogLeNet %http://www.programmersought.com/article/7609143938/ we add a new randomly initialized fully connected layer with $128$ nodes, following by a $l2$ normalization which will be used as the feature embeddings. 
The embedding module (F) is added after the last average pooling of the GoogLeNet. 
 A softmax layer is used as the clustering module to output a probabilistic vector for each image embedding. 
The decoder (D) is a stack of deconvolutional layers in which each layer is followed by a batch norm layer and ReLU activation. 
%When training the whole framework, we first pre-train the framework with the reconstruction loss only. More precisely, because we do not want the encoder to overfit the reconstruction criterion, we freeze the encoder part (i.e., the GoogleNet part) and only train the decoder to get the initialization for the decoder (D). After that, we train the whole network with the final loss as in \eqref{eq:final_loss}. 
The values of $\alpha$, $\beta$, and $\gamma$ in \eqref{eq:final_loss} are empirically set to 0.9, 0.3, 0.01, respectively. We adopt the SGD optimizer with momentum {0.9}. 
To make a fair comparison to SME~\cite{ye2019unsupervised}, similar to that work, we also use the batchsize 64 (i.e., 128 after augmentation). 
%~\footnote{\gustavo{This small batchsize for SME gives better performance than larger batchsize. We believe that is because of the increase in the number of false negative samples that is probably happening in larger batchsizes. We actually run SME (using the released code) with batchsize 256 and R@1 on CUB is 45.8, which is lower than the 46.2 reported in SME~\cite{ye2019unsupervised}.}}
 The value of the temperature $\tau$ in (\ref{eq:close}), (\ref{eq:far}) is set to 0.1.
%The initial learning rate for all layers but the decoder part is set to {0.001}.
}
{%During training, all input images are firstly resized to $256\times 256$. 
For the data augmentation to create positive samples, the original images are randomly cropped at size $224\times 224$ with random horizontal flipping which is similar to \cite{iscen2018mining,ye2019unsupervised}. In the testing phase, as a standard practice \cite{iscen2018mining}, a single center-cropped image is used as input to extract the image embedding. 
}

\subsection{Dataset and Evaluation Metric}
{We conduct our experiments on two public benchmark datasets that are commonly used to evaluate DML methods, where we follow the standard experimental protocol for both datasets~\cite{smart,npair,clustering,lifted}. 
The \textbf{CUB200-2011} dataset~\cite{cub} contains 200 species of birds with 11,788 images, where the first 100 species with 5,864 images are used for training and the remaining 100 species with 5,924 images are used for testing. The \textbf{Car196} dataset~\cite{car} contains 196 car classes with 16,185 images, where the first 98 classes with 8,054 images are used for training and the remaining 98 classes with 8,131 images are used for testing. We report the K nearest neighbor retrieval accuracy using the Recall@K metric. We also report the clustering quality using the normalized mutual information (NMI) score~\cite{nmi}.  
}

\subsection{Ablation Study}

{In this section, we investigate the impact of each loss term in the proposed method. We also investigate the effect of the number of centroids in the clustering module which affects the pseudo labels. Here we consider the work SoftMax Embedding (SME) \cite{ye2019unsupervised} as the baseline model. %, i.e., given an anchor point, its augmented version is defined as the positive sample, and all other samples in the batch are considered as the negative samples. SME's metric loss has similar form to our metric loss (\ref{eq:ml_loss}). However, in \cite{ye2019unsupervised} the loss is calculated using each anchor point, its positive and negative samples. In our work, we calculate the metric loss using anchor points and centroids. 
We denote our model that is trained by using only clustering loss (\ref{eq:c_loss}) as \textbf{only $L_{rim}$} and our model that is trained with both clustering and the metric losses (\ref{eq:c_loss}) and (\ref{eq:ml_loss}) as \textbf{Center-based Softmax (CBS)}. In this experiment, %the number of clusters are fixed to $150$ and $75$ for CUB200-2011 and Car196, respectively. 
the number of clusters are fixed to $32$ for both CUB200-2011 and Car196 datasets. 
We also investigate the impact of the reconstruction in enhancing the representativeness of centroids. This model is denoted as \textbf{Center-based Softmax with Reconstruction (CBSwR)}, which is our final model. 

Tables \ref{tab:ablation_cub200} and \ref{tab:ablation_Car196} present the comparative results between methods. The results show that using only the clustering loss, the accuracy is significantly lower than the baseline. However, when using the centroids from the clustering for calculating the metric loss (i.e., CBS), it gives the performance boost over the baseline (i.e., SME). %This confirms the effectiveness of our proposal, i.e., using the centroids for more accurate negative mining. 
Furthermore, the reconstruction loss enhances the representativeness of centroids, as confirmed by the improvements of CBSwR over CBS on both datasets. 
%\gustavo{This ablation study shows the impact of each loss component and their combinations, and results show that each loss complements each other and helps to improve the performance.}

Table \ref{tab:training_time} presents the training time of different methods on the CUB200-2011 and Car196 datasets. Although the asymptotic complexity of CBSwR for training one batch is $\mathcal{O}(Km)$, it also consists of a decoder part which affects the real training. It is worth noting that the decoder is only involved during training. During testing, our method has similar computational complexity as SME.

Table \ref{tab:num_clusters} presents the impact of the number of clusters $K$ in the clustering loss on the CUB200-2011 dataset with our proposed model CBSwR (recall that the number of clusters $K$ is limited by the batch size $m$). %During training, there was no empty cluster in all experiments.
During training, the number of samples per clusters vary depending on batches and the number of clusters. At $K=32$ which is our final setting, the number of samples per cluster varies from 2 to 11, on the average.
%During training of batches, there may be some empty clusters (especially when $K$ is large). In those cases, we simple ignore empty clusters and calculate the metric loss as normal.
%\textcolor{red}{The number of empty clusters when testing is 0 in all experiments.}
%Interestingly, although there are 100 classes in CUB200-2011 dataset, we found that using $100$ clusters does not make any significant difference in the retrieval accuracy. 
The retrieval performance is just slightly different for the different number of clusters. This confirms the robustness of the proposed method w.r.t. the number of clusters. %the number of true semantic classes in datasets which is not always available. 

%Figure \ref{fig:reconstruction} displays samples from CUB200-2011 dataset and their reconstructed images using the corresponding centroids of the clusters that they belong. We can see that most samples belonging to the same class (Figure \ref{fig:CUBsamples}) are accurately clustered into the same cluster (Figure \ref{fig:reconstructedCUB}), i.e., they have the same reconstructed image. Furthermore, the reconstructed images using the centroids  show an ``average image'' that is visually similar to input samples for each class. This further confirms the representativeness of centroids \gustavo{mentioned in Sec.~\ref{sec3:r_loss}}. 
}

\begin{table}[!t]
\centering
\small
\begin{center}
\begin{tabular}{c |m{1.2cm} m{1.2cm} m{1.2cm} m{1.2cm}}
\hline
& {R@1}  & {R@2}  & {R@4}  & {R@8} %&\textbf{NMI}  
                   \\ \hline
SME \cite{ye2019unsupervised}    & 46.2  & 59.0          & 70.1          & 80.2          %& 55.4          
\\
only $L_{rim}$    & 40.6  & 52.9          & 65.7          & 77.5         
\\ 
CBS         & 47.3          & 59.1     & 70.5          & 80.2         %& 55.3    
\\ 
%CBSwR & {48.5} & {60.0} & {71.3} & {81.1} %& \textbf{55.4} 
% CBSwR & {48.1} & {59.4} & {70.8} & {81.0} 
CBSwR & {47.5} & {59.6} & {70.6} & {80.5} 
\\ \hline
\end{tabular}
\end{center}
\caption{\small The impact of each loss component on the performance on CUB200-2011 dataset and the comparison to the baseline \cite{ye2019unsupervised}.}
\label{tab:ablation_cub200}
\end{table}

\begin{table}[!t]
\centering
\small
\begin{center}
\begin{tabular}{c|  m{1.2cm} m{1.2cm} m{1.2cm} m{1.2cm}}
\hline
  & {R@1}  & {R@2}  & {R@4}  & {R@8}  %& \textbf{NMI}  
  \\ \hline
SME \cite{ye2019unsupervised}    & 41.3& 52.3   & 63.6 & 74.9       %& 35.8      
\\ 
only $L_{rim}$  & 35.7  & 47.5  & 59.8 & 71.0
\\
CBS      & 42.2 & 54.0  & 65.4  & 76.0%& 36.8
\\ 
% CBSwR & {43.1} & {54.7} & {65.7} & {76.0}  %&\textbf{37.2} 
CBSwR & {42.6} & {54.4} & {65.4} & {76.0}
\\ \hline
\end{tabular}
\end{center}
\caption{\small The impact of each loss component on the performance on Car196 dataset and the comparison to the baseline \cite{ye2019unsupervised}.}
\label{tab:ablation_Car196}
\end{table}

\begin{table}[!t]
\centering
\small
\begin{center}
\begin{tabular}{ c |m{1.2cm} m{1.2cm} m{1.2cm} m{1.2cm} }
\hline
{Dataset} & {SME} & {only $L_{rim}$} & {CBS} & {CBSwR} %& \textbf{NMI} 
\\ \hline
CUB200-2011 &696 &620 &660 & 882
\\ 
Car196 & 772 & 700 & 727 & 1025
\\ \hline
\end{tabular}
\end{center}
\caption{\small The training time (seconds) of different methods on CUB200-2011 and Car196 datasets with 20 epochs. The models are trained on a NVIDIA GeForce GTX 1080-Ti GPU.}
\label{tab:training_time}
\end{table}

\begin{table}[!t]
\centering
\small
\begin{center}
\begin{tabular}{ c |m{1.2cm} m{1.2cm} m{1.2cm} m{1.2cm} }
\hline
{Number of clusters} & {R@1} & {R@2} & {R@4} & {R@8} %& \textbf{NMI} 
\\ \hline
16 &46.7 &59.0 &70.0 & 79.7
\\ 
32 & 47.5 & 59.6 & 70.6 & 80.5 %& 55.1         
\\ 
48& 47.1& 58.9         & 70.6         & 80.1         %& 55.4
\\ 
64                      & 47.4         & 58.9         & 69.6         & 80.2%& 56.0 
\\ \hline
\end{tabular}
\end{center}
\caption{\small The impact of the number of clusters of the final model CBSwR on the performance on CUB200-2011 dataset.}
\label{tab:num_clusters}
\end{table}

% \begin{figure*}[!t]
% \centering
% \subfigure[CUB200-2011 samples]{
%       \includegraphics[scale=0.15]{figure/cub-sample.png}
%       \label{fig:CUBsamples}
% }
% \subfigure[Reconstructed images]{
%       \includegraphics[scale=0.15]{figure/cub-reconstruction.png} 
%       \label{fig:reconstructedCUB}
% }
% \caption[]{a) sample images from CUB200-2011 dataset. 
% Box colors represent for class labels. b) reconstructed images using corresponding centroids. Box colors denote for cluster indexes. 
% }
% \label{fig:reconstruction}
% \end{figure*}

\begin{table*}[!t]
	\centering
	\small
	\begin{center}
		\begin{tabular}{m{4cm} m{1.2cm} m{1.2cm} m{1.2cm} m{1.2cm} m{1.2cm}} 
			\hline
			&NMI &R@1 &R@2 &R@4 &R@8 \\ \hline
	&\multicolumn{5}{c}{Supervised Learning}
			\\ \hline
			SoftMax &57.2 &48.3 &60.2 &71.2 &80.3 \\
			%Semi-hard~\cite{facenet} &55.4 &42.6 &55.0 &66.4 &77.2 \\
			%Lifted structure~\cite{lifted} &56.5 &43.6 &56.6 &68.6 &79.6 \\
			N-pair~\cite{npair} &57.2 &45.4 &58.4 &69.5 &79.5 \\
			%Triplet+Global~\cite{global} &58.6 &49.0 &61.0 &72.3 &81.9 \\
			%Clustering~\cite{clustering} &59.2 &48.2 &61.4 &71.8 &81.9 \\
			Triplet+smart mining~\cite{smart} &59.9 &49.8 &62.3 &74.1 &{83.3} \\
			Triplet+proxy~\cite{tripletproxy} &59.5 &49.2 &61.9 &67.9 &72.4 \\
			Histogram~\cite{histogram} &- &50.3 &61.9 &72.6 &82.4 \\
			%Discriminative \cite{do2019theoretically} &\textbf{59.92} &\textbf{51.43} &\textbf{64.23} &\textbf{74.31} &{82.83} \\
			\hline
			&\multicolumn{5}{c}{Unsupervised Learning} \\ \hline
Cyclic \cite{DBLP:conf/eccv/LiHHWA016} &52.6  &40.8  &52.8 &65.1 &76.0  \\ 
Exemplar \cite{dosovitskiy2015discriminative}&45.0 &38.2 &50.3 &62.8 &75.0   \\ 
NCE \cite{wu2018unsupervised}        &45.1 &39.2 &51.4 &63.7 &75.8 \\ 
DeepCluster \cite{caron2018deep}&53.0 &42.9 & 54.1 & 65.6 &76.2 \\ 
MOM  \cite{iscen2018mining}          &55.0 &45.3 &57.8  &68.6&78.4\\ 
SME \cite{ye2019unsupervised}      &55.4 &46.2 &59.0 &70.1 &80.2 \\ 
CBSwR (Ours)    &\textbf{55.9} & \textbf{47.5}&\textbf{59.6}     &\textbf{70.6} &\textbf{80.5} \\ \hline
		\end{tabular}
	\end{center}
	\caption{\small Clustering and Recall performance on the CUB200-2011 dataset.} 
	\label{tab:cub}
\end{table*}

\begin{table*}[!t]
	\centering
	\small
	\begin{center}
		\begin{tabular}{m{4cm} m{1.2cm} m{1.2cm} m{1.2cm} m{1.2cm} m{1.2cm}} 
			\hline
			&NMI &R@1 &R@2 &R@4 &R@8 \\ \hline
				&\multicolumn{5}{c}{Supervised Learning}
			\\ \hline
			SoftMax &58.4 &62.4 &73.0 &80.9 &87.4 \\
			%Semi-hard~\cite{facenet} &53.4 &51.5 &63.8 &73.5 &82.4 \\
			%Lifted structure~\cite{lifted} &56.9 &53.0 &65.7 &76.0 &84.3 \\
			N-pair~\cite{npair} &57.8 &53.9 &66.8 &77.8 &86.4 \\
			%Triplet+Global~\cite{global} &58.2 &61.4 &72.5 &81.8 &88.4 \\
			%Clustering~\cite{clustering} &59.0 &58.1 &70.6 &80.3 &87.8 \\
			Triplet+smart mining~\cite{smart} &59.5 &64.7 &76.2 &84.2 &90.2 \\
			Triplet+proxy~\cite{tripletproxy} &{64.9} &{73.2} &{82.4} &{86.4} &88.7 \\
			Histogram~\cite{histogram} &- &54.3 &66.7 &77.2 &85.2 \\
			%Discriminative \cite{do2019theoretically}&59.71 &68.31 &78.21 &85.22 &\textbf{91.18} \\ 
			\hline
			&\multicolumn{5}{c}{Unsupervised Learning}\\ \hline
Exemplar \cite{dosovitskiy2015discriminative}&35.4 &36.5 &48.1 &60.0 &71.0 \\
NCE \cite{wu2018unsupervised}  &35.6 & 37.5 &48.7 &59.8 &71.5  \\ 
DeepCluster \cite{caron2018deep}&38.5 &32.6 &43.8&57.0 &69.5 \\ 
MOM  \cite{iscen2018mining} &\textbf{38.6} &35.5 &48.2&60.6&72.4  \\ 
SME \cite{ye2019unsupervised}&35.8 &41.3&52.3 &63.6&74.9 \\ 
CBSwR (Ours) &37.6&\textbf{42.6} &\textbf{54.4} 
&\textbf{65.4}&\textbf{76.0}       \\\hline
		\end{tabular}
	\end{center}
	\caption{\small Clustering and Recall performance on the Car196 dataset.} 
	\label{tab:car}
\end{table*}

% \begin{figure*}[!t]
%     \centering
%     \includegraphics[width=0.90\textwidth, keepaspectratio=true]{figure/cub200_visualization.jpg}
%     \caption{Barnes-Hut t-SNE visualization \cite{van2014accelerating} of our embedding on the %test split of 
%     CUB200-2011 dataset. Best view in color.}
%     \label{fig:cub200_visualization}
% \end{figure*}

\subsection{Comparison to the State of the Art}
{
We compare our method to the state-of-the-art unsupervised deep metric learning and unsupervised feature learning methods that have reported results on the benchmarking datasets CUB200-2011 and CARS196. They are the graph-based method \textbf{Cyclic} \cite{DBLP:conf/eccv/LiHHWA016}, Exemplar CNN with surrogate classes from image transformations \textbf{Exemplar} \cite{dosovitskiy2015discriminative}, feature learning with noise-contrastive estimation \textbf{NCE} \cite{wu2018unsupervised},  alternative feature learning and k-means clustering \textbf{DeepCluster} \cite{caron2018deep}, mining on manifold \textbf{MOM}  \cite{iscen2018mining}  and \textbf{SoftmaxEmbedding --- SME} \cite{ye2019unsupervised}. Among them, MOM \cite{iscen2018mining} and SME  \cite{ye2019unsupervised} are the only methods that claim for unsupervised deep metric learning. 
We also compare our method to fully supervised deep metric learning methods, including the \textbf{softmax} loss, %the \textbf{triplet} loss with \textbf {semi-hard} negative mining~\cite{facenet}, the \textbf{lifted structured} loss~\cite{lifted}, 
the \textbf{N-pair}~\cite{npair} loss, 
%the \textbf{clustering} loss~\cite{clustering}, the \textbf{triplet} combined with \textbf{global} loss~\cite{global}, 
the \textbf{histogram} loss~\cite{histogram}, the \textbf{triplet with proxies}~\cite{tripletproxy} loss, \textbf{triplet with smart mining} ~\cite{smart} loss which uses the fast nearest neighbor search for mining triplets. 

Table \ref{tab:cub} presents the comparative results on CUB200-2011 dataset. In terms of clustering quality (i.e., NMI metric), the proposed method and the state-of-the-art UDML methods MOM \cite{iscen2018mining} and SME \cite{ye2019unsupervised} achieve comparable accuracy. However, in terms of retrieval accuracy R@K, our method outperforms other approaches. %especially at R@1, i.e., we outperform SME \cite{ye2019unsupervised} by $2.3\%$. 
Our proposed method is also competitive to most of the supervised DML methods. %We are aware that there are supervised methods, e.g. Discriminative \cite{do2019theoretically}, that achieve better performance than ours. It is understandable because those methods use the label information for exactly selecting positive and negative samples during training. 

Table \ref{tab:car} presents the comparative results on Car196 dataset. Compared to unsupervised methods, the proposed method  outperforms other approaches in terms of retrieval accuracy at all ranks of $K$. %, i.e., our method outperforms the state-of-the-art UDML methods SME by around $2\%$ at R@K=1,2,4. 
Our method is comparable to other unsupervised methods in terms of clustering quality. %We also found that there are substantial differences between supervised and unsupervised methods in both Recall@K and NMI scores on Car196. One possible reason is that Car196 is more challenging than CUB200-2011, where the visual information present in images from different classes are less discriminative than those in CUB200-2011. 
%Hence, an unsupervised mining of negative samples is less accurate compared to CUB200-2011. In that case, a strong supervision information, such as label information, is required to improve the mining and retrieval performances.

%Figure \ref{fig:cub200_visualization} shows the t-SNE \cite{van2014accelerating} plots on our learned embedding features on CUB200-2011. We can see that our embedding produces reasonable results in grouping similar visual objects despite the significant variations in view point, pose, and configuration. 
}

\section{Conclusion}
{
We propose a new method that utilizes deep clustering for deep metric learning to address the two challenges in UDML, i.e., positive/negative mining and efficient training. %Our method is based on a novel unsupervised deep metric learning approach for optimizing deep metric embedding with a learnable clustering function, a reconstruction function, and a center-based metric loss function in an end-to-end fashion using a principled structured prediction framework.
The method is based on a novel loss that consists of  a learnable clustering function, a reconstruction function, and a center-based metric loss function. 
Our experiments on CUB200-2011 and Car196 datasets show state-of-the-art performance on the retrieval task, compared to other unsupervised learning methods. 
}

\bibliography{egbib}
\end{document}